# RM-PoT: Reformulating Mathematical Problems and Solving via Program of Thoughts

Yu Zhang$^{\S\flat}$, Shujun Peng$^{\S}$, Nengwu Wu$^{\triangle}$, Xinhan Lin$^{\flat}$, Yang Hu$^{\S}$, Jie Tang$^{\S}$

$\S$Tsinghua University, China; $\flat$Shanghai Artificial Intelligence Laboratory, China;
$\triangle$University of Electronic Science and Technology, China

## ABSTRACT

Recently, substantial advancements have been made in training language models to carry out step-by-step reasoning for solving intricate numerical reasoning tasks. Beyond the methods used to solve these problems, the structure and formulation of the problems themselves also play a crucial role in determining the performance of large language models. We observe that even small changes in the surface form of mathematical problems can have a profound impact on both the answer distribution and solve rate. This highlights the vulnerability of LLMs to surface-level variations, revealing its limited robustness when reasoning through complex problems. In this paper, we propose RM-PoT, a three-stage framework that integrates problem reformulation (RM), code-aided reasoning (PoT), and domain-aware few-shot learning to address these limitations. Our approach first reformulates the input problem into diverse surface forms to reduce structural bias, then retrieves five semantically aligned examples from a pre-constructed domain-specific question bank to provide contextual guidance, and finally generates executable Python code for precise computation.

## 1 INTRODUCTION

Mathematical reasoning is a cornerstone of problem-solving, with applications spanning diverse fields such as physics, engineering, economics, and computer science. However, despite their success in general natural language processing tasks, existing Large Language Models (LLMs) such as GPT-4 struggle with mathematical problems that demand precision, logical reasoning, and step-by-step computation Hendrycks et al. (2021). This gap arises because LLMs rely heavily on statistical patterns in natural language, which often fail to capture the formal structure and symbolic complexity of mathematical problems. Cobbe et al. (2021).

Current advancements in mathematical reasoning with LLMs have primarily focused on methods like Chain-of-Thought (CoT) prompting Wei et al. (2022); Diao et al. (2023); Zhang et al. (2022), which encourages models to break problems into reasoning steps. Self-Consistency (SC) methods further refine CoT by introducing multiple reasoning paths to identify consistent answers. While effective, these approaches are still limited when faced with problems requiring complex computation or diverse logical forms. As shown in Fig 1(a), the LLM fails to provide the correct answer directly when posed with a complex calculation problem.

This limitation is handled by another notable approach, Program of Thoughts (PoT), which introduces intermediate code generation to represent reasoning steps explicitly, improving the interpretability of solutions and disentagle computation from the reasoning process. However, PoT alone cannot resolve inconsistencies arising from ambiguous structured problem forms. As shown in Fig 1(b), we can observe that small changes in the surface form of the mathematical problem can lead to significantly different outcomes.

To address these challenges, we propose RM-PoT, a novel two-stage framework that integrates Reformulation of Mathematical Problems (RM) and Program of Thoughts (PoT). The RM stage prompts the LLM to generate multiple paraphrased versions of the same problem, thereby mitigating the adverse effects of poorly structured problem formulations. By generating multiple reformulations of a given problem and exposing the model to various formulations of the same task, LLMs can uncover the core mathematical structure underlying the problem zho. This diversity in problem presentation improves the LLM's accuracy and consistency in solving the problem. In the PoT stage, the model generates Python code as intermediate reasoning steps, decoupling symbolic computations from natural language reasoning. A voting mechanism ensures consistency across reformulated solutions, enhancing accuracy and robustness. Domain-specific context activation can retrieve semantically coherent ⟨question-solution⟩ pairs from a pre-built question bank, enabling LLM to reason based on specific domains.

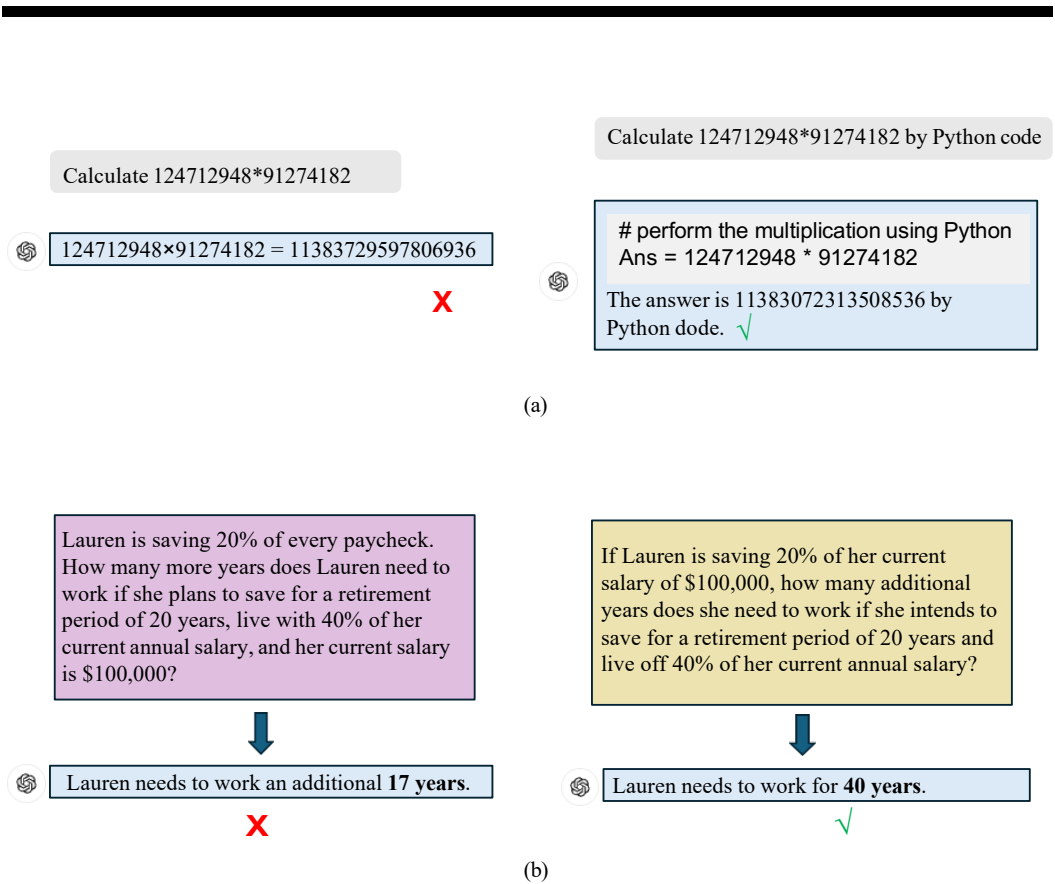

Figure 1: Limitations for GPT-4 solving MWPs. (a) When GPT-4 is asked to directly provide the result of a complex arithmetic problem, the answer is incorrect.(b) When the original problem is reformulated, answers are inconsistent.

During the deduction phase, we introduce a novel integration of Few-Shot Learning module to further improve the accuracy and robustness of LLMs in mathematical problem-solving which operates as follows : Given a problem p, we (1) classify its mathematical domain D (e.g., algebra, arithmetic) using a lightweight embedding model, (2) retrieve the top-5 ⟨question,solution⟩ pairs from D's bank via cosine similarity over sentence-BERT embeddings, and (3) inject these examples into the prompt to prime the LLM with domain-specific reasoning patterns. This process ensures semantic alignment between the input problem and the few-shot exemplars, enhancing both reformulation diversity and solution accuracy. Additionally, we analyze the performance gains from the inclusion of few-shot examples, showing that incorporating these examples allows for more accurate model predictions, even when faced with unfamiliar or highly variable problem structures.

The main contributions of this paper are as follows:

- We introduce RM-PoT, a novel framework that combines surface-level problem reformulation and explicit intermediate code generation to improve mathematical reasoning in LLMs.
- We conduct comprehensive experiments on widely-used benchmarks, including GSM8K, AQuA, and MATH, demonstrating that RM-PoT outperforms baseline approaches across different datasets.
- We provide analysis of the effectiveness of RM and PoT stages, highlighting their contributions to solving mathematical problems.

## 2 RELATED WORK

### 2.1 MATHEMATICAL REASONING IN LLMS

In recent years, numerous studies have explored using the System-2 reasoning approach to solve mathematical problems with LLMs Wei et al. (2022); **?**); **?**. As a prominent framework, chain-of-thought(CoT) is proposed by Wei et al. (2022). Rather than generate the answer directly, it prompts the LLMs to produce a sequence of intermediate reasoning steps. **?** further extended CoT by self-consistency. They generate multiple reasoning steps from different angles and potentially lead to the same answer. Chen et al. (2022) proposes program-of-thoughts, using intermediate codes to represent the reasoning process.

### 2.2 SPECIALIZED MODELS FOR MATHEMATICAL TASKS

To address the limitations of general-purpose LLMs, specialized approaches have been developed. For instance, MathGPT **?** integrates symbolic computation engines to solve algebraic problems, combining language modeling with symbolic reasoning. However, this hybrid approach still depends heavily on the capabilities of external tools. Other models, such as GeoSolver **?** and MathQA **?**, focus on specific domains like geometry or math question-answering, using domain-specific datasets and tailored architectures. While these models perform well in their respective areas, their narrow focus limits generalization to broader mathematical tasks.

### 2.3 PROBLEM REFORMULATION IN MATHEMATICAL PROBLEM SOLVING

The role of problem reformulation in enhancing the performance of language learning models (LLMs) in mathematical problem solving has gained increasing attention in recent years. Early work by **??** demonstrated that altering the structure and phrasing of a problem can lead to improved reasoning outcomes in LLMs. This line of research has been extended by **?**, who explored how different surface forms of mathematical problems can significantly influence the solve rate of LLMs. Further studies have investigated the impact of reformulation strategies such as paraphrasing **?** and introducing problem variants **?**, suggesting that diverse representations help models better understand and process complex tasks. Building on these insights, recent works have also delved into the combination of reformulation with in-context learning techniques to optimize model performance through adaptive problem presentations.

## 3 METHOD

We propose RM-POT, a framework that leverages the potential of large language models (LLMs) to solve mathematical problems. The overall architecture of RM-POT is illustrated in Fig. 2, consisting of two stages.

- **Stage I**: We reformulate the given mathematical problems into diverse surface forms, enabling the LLM to better grasp the underlying structure of the problems.
- **Stage II**: We employ the Program of Thoughts Chen et al. (2022) (PoT) approach, which generates Python code to solve the reformulated problems. This approach not only reveals the reasoning process of the LLM but also separates the computational steps from the reasoning process.

The details of these two stages are described in the following subsections.

### 3.1 REFORMULATE MATHEMATICAL PROBLEMS(RM)

As demonstrated in Section 1, the surface form of a problem significantly influences the performance of LLMs zho. Therefore, we prompt the LLM to generate $K$ different surface forms of the original problem and use a voting mechanism to determine the answer. The intuition behind this approach is that if a problem exhibits a low solve rate and ineffective reasoning paths due to its original surface form, introducing diversity in its surface forms can enhance the likelihood of finding a correct solution.

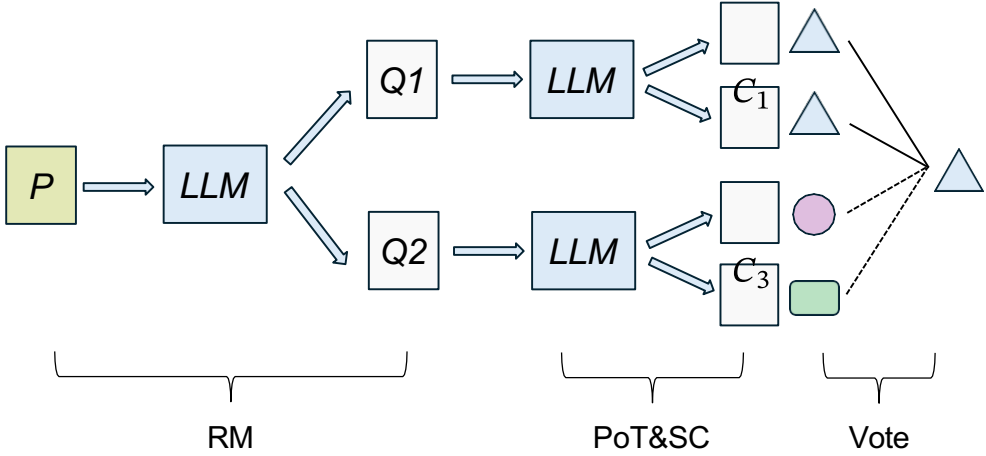

Figure 2: Overview of our proposed RM-PoT framework. It operates in two stages. First, the RM method is employed to reformulate the original problem into various forms using the LLM. Next, the PoT approach is utilized to guide the LLM in generating intermediate code, which not only illustrates the reasoning process but also computes the result. The final answer is then determined through a voting mechanism.

It is important to note that the original problem is reformulated into different forms using the same LLM that is used to solve the problem. This ensures that the improvement in model performance is due to the diversity of the reformulated problem forms, rather than the sharing of knowledge with other LLMs.

In this paper, we explore two ways of prompting LLM to generate reformulated problems. The naive prompt template is "*Reformulate the following math problem, try to change the sentence structure of the problem*:{input problem}" . The In-Context method begins by identifying effective examples: We conduct experiments to determine which pairs of (Original Problem, Reformulated Problem) exhibit the largest margin in solve rates. Then, we use these examples to enable in-context learning for the LLM, as depicted in Fig. 3.

### 3.2 PROGRAM OF THOUGHTS(PoT)

Program of Thoughts (PoT) is a method aimed at enhancing the reasoning capabilities of large language models (LLMs) Chen et al. (2022). It works by decomposing complex tasks into a series of intermediate steps, or "thoughts", which guide the model through a structured reasoning process. Each thought represents a logical step that incrementally leads to the final answer, enabling the model to tackle intricate problems using a step-by-step approach.

This method improves the model's ability to solve tasks that require multi-step reasoning, such as mathematical or logical problems, by fostering transparency in the reasoning process and increasing accuracy in the final result. PoT has proven effective in scenarios where direct answers are difficult, allowing LLMs to perform more reliably in problem-solving tasks.

In our proposed RM-PoT, we aim to instruct the LLM to generate intermediate Python code to solve mathematical problems. This approach can clearly show the reasoning process of LLM and improve accuracy.

### 3.3 SELF-CONSISTENCY(SC) AND VOTING

In our proposed RM-PoT framework, we reformulate the original problems into $K$ different surface forms. Morever, we utilize the Self-Consistency(SC) **?** method. Specifically, for each formulated problem, we let LLM generate $\frac{K}{N}$ reasoning paths, and thus the total number of generated answer is $N$. We then vote for the final answer.

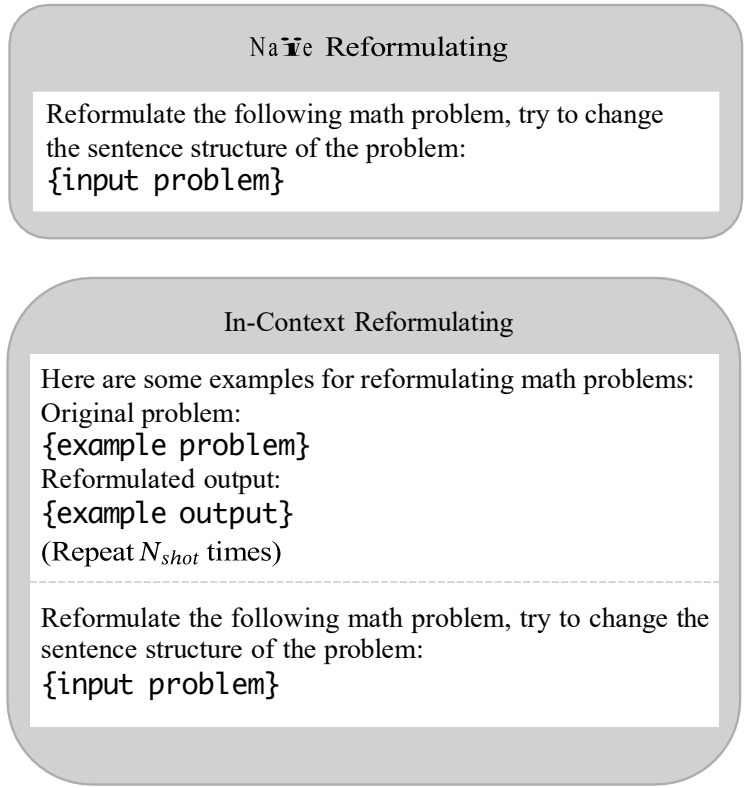

Figure 3: Naive and In-Context Reformulating prompts for LLM.

By increasing $K$, we maintain a fixed total number of reasoning paths $N$. This approach effectively isolates the effect of diversifying the reasoning paths from merely increasing their number. It also ensures a fair comparison with the self-consistency baselines.

## 4 EXPERIMENTS

### 4.1 EXPERIMENTAL SETTINGS

**Datasets** We evaluate our approach on the following public mathematics reasoning benchmarks:

- **GSM8k** Cobbe et al. (2021) contains 8.5K linguistically diverse grade school-level math questions with moderate difficulties.
- **AQuA ?** consists of 100K algebraic word problems, including the questions, the possible multiple-choice options, and natural language answer rationales from GMAT and GRE.
- **SVAMP ?** contains 1K arithmetic word problems. It focuses on basic arithmetic operations such as addition, subtraction, multiplication, and division.

**Large Language Model** We use GLM-4-9B ? as our base model. All experiments are conducted in zero-shot or few-shot settings, without training or fine-tuning it. For generation configs, We set the temperature $T = 0.7$, Top-p= 0.8 and Top-k= 3. The total number of reasoning paths $N$ we sample for each problem is 16.

**Implementation Details** For problem solving, the PoT process consists of two steps, as illustrated in Fig. 4. First, we prompt the LLM to generate Python code and store the result in a variable with a fixed name, allowing for convenient extraction. If the problem includes options, we instruct the LLM to identify the closest match among the provided options.

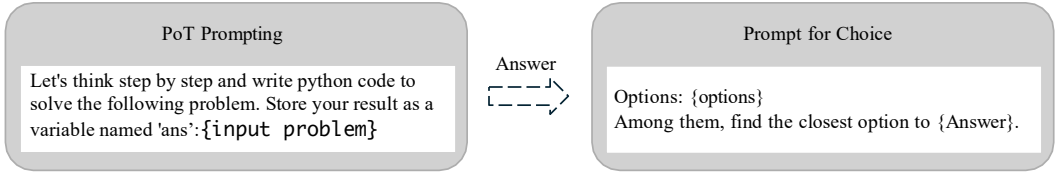

Figure 4: Prompts for solving the problem and selecting the correct choice when options are provided.

## 4.2 EFFECTIVENESS OF REFORMULATING

To verify the effectiveness of RM, we reformulate the selected problems from the AQuA dataset and calculate the solve rate difference between the original problems and their reformulated versions. An example is shown in Fig. 5 to provide readers with an intuitive understanding of the RM process. When the surface form of the original problem is altered, the solve rate increases from 43.8% to 81.3%.

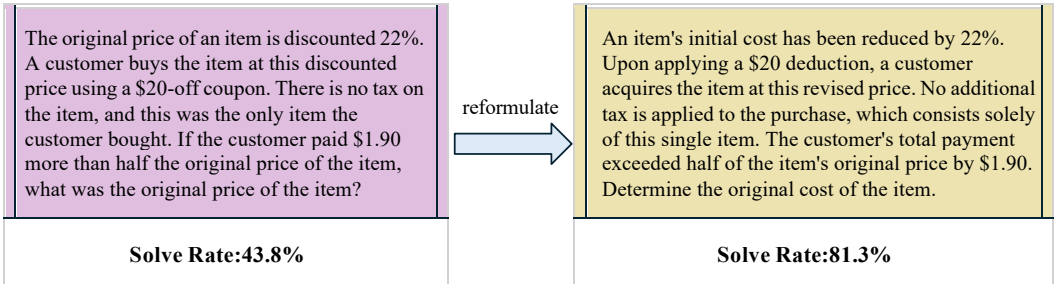

Figure 5: An example from AQuA dataset. When the original problem is reformulated, the solve rate improves significantly.

Furthermore, we compute the solve rate difference between the original problems $p$ and the reformulated problems $p_r$ from the AQuA dataset, defined as where $SR(p_r) - SR(p)$, where $SR$ denotes the solve rate. The results are shown in Fig. 6. we can observe that reformulating the problems leads to an overall improvement in the solve rate.

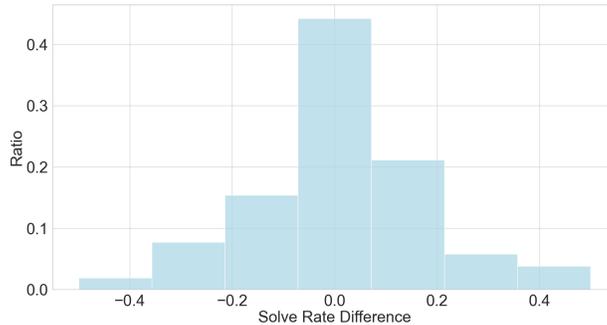

Figure 6: The solve rate difference between the original problems and the reformulated versions.

## 4.3 MAIN RESULTS

In this subsection, we compare the performance of RM-PoT with Chain of Thoughts (CoT), vanilla self-consistency (SC), and vanilla Program of Thoughts(PoT). All results are presented in Table 1. In this evaluation, we apply the naive reformulation of the original problems. the number of reformulated problems is set to $K = 4$, and the total number of reasoning paths is $N = 16$. For SC, we also maintain $N = 16$.

The performance metric presented in the table is the accuracy of answers after voting. We can observe that RM-PoT outperforms the other three baselines across all datasets, despite the considerable likelihood of generating reformulation with a lower solve rate (see Fig. 6). We assume that this may be because the model demonstrates more consistent understanding of the problems when exposed to various reformulations, thereby avoiding misinterpretations of the original problem in certain specific cases. We will further discuss it in section 5.

Table 1: Comparison of accuracy between different methods.

|        | GSM8K | AQuA | SVAMP |
|--------|-------|------|-------|
| CoT    | 75.6  | 63.2 | 86.3  |
| SC     | 77.3  | 64.9 | 87.6  |
| PoT    | 78.9  | 65.6 | 87.1  |
| RM-PoT | **80.4** | **69.1** | **88.0** |

### 4.4 ABLATION STUDY

In this subsection, we vary the number of reformulated problems $K$ across $\{1,2,4\}$, while keeping the total reasoning paths fixed at $N = 16$. Additionally, we compare the naive reformulation with the In-Context reformulation, as outlined in Fig 3. The results are shown in Table 2.

Table 2: Comparison of accuracy between different settings of $K$ and reformulation.

|            | K | GSM8K | AQuA | SVAMP |
|------------|---|-------|------|-------|
|            | 1 | 78.2  | 66.0 | 86.9  |
| Naive      | 2 | 79.8  | 67.7 | **88.4** |
|            | 4 | **80.4** | **69.1** | 88.0 |
|            | 1 | 78.4  | 67.6 | 87.1  |
| In-Context | 2 | 80.1  | 69.4 | 89.0  |
|            | 4 | **80.9** | **72.2** | **89.6** |

We observe that as $K$ increases, the performance of the LLM improves as well. This further validates our assumption that exposing the LLM to different surface forms of a problem allows it to better grasp the underlying structure. Notably, on the SVAMP dataset, the LLM performs better when $K = 2$ than when $K = 4$. This may be because the problems in this dataset are relatively simple, and the process of reformulating and solving the problems involves some degree of randomness. By comparing the naive reformulation with the In-Context reformulation, we can conclude that the LLM learns the reformulation method more effectively when provided with good examples.

## 5 DISCUSSION

In this section, we discuss how RM-PoT helps the LLM solve math problems through the example shown in Fig. 7. When applying CoT, the LLM makes two mistakes. First, there is an error in rearranging the terms of the equation in the second step.econd, the result of 21.90/1.60 is actually 20.660, not 20.741. This demonstrates that the LLMs has certain shortcomings in computation and reasoning.

To disentangle computaion from reasoning process, we then apply the PoT method. We find that the LLM's answer is generally correct, but it misinterprets the original price as the discounted price during the understanding of the problem. However, after reformulating the original problem, the LLM correctly understands the question and provides the correct answer. Although both versions of the problem contain the term 'original price,' the LLM does not fully grasp this in the original formulation. The deeper reasons behind this remain unclear and will be explored in future work.

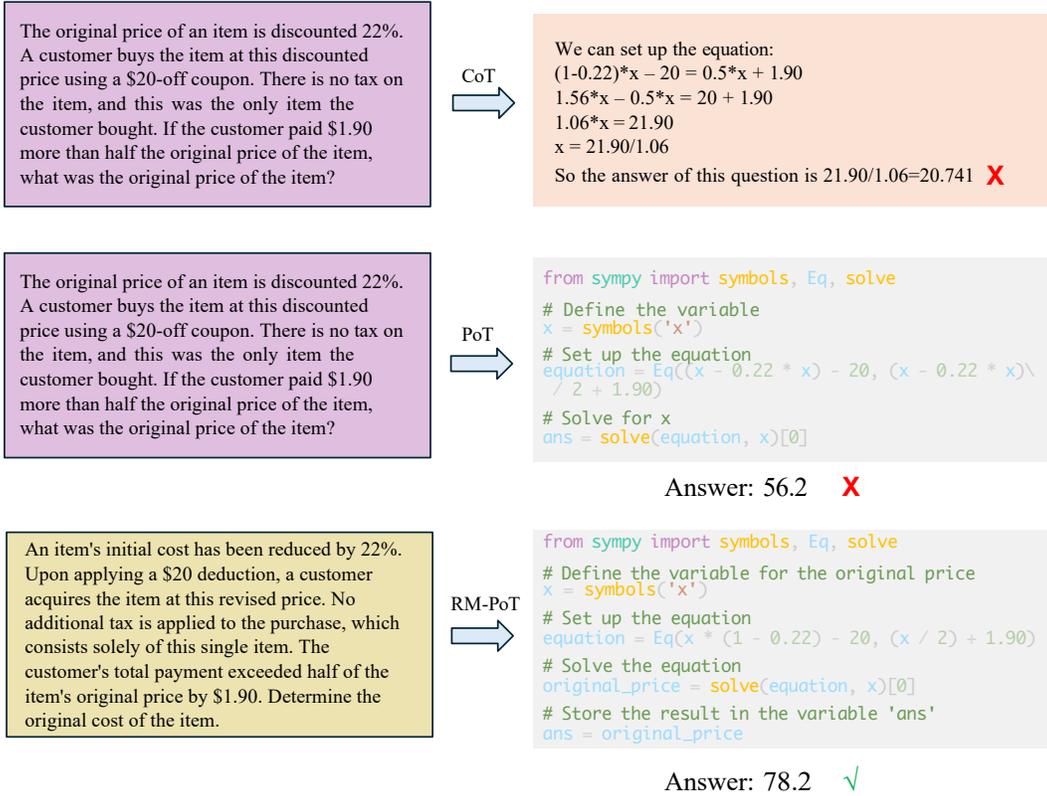

Figure 7: An example showing how CoT, PoT, and RM-PoT solve the same problem. Both CoT and PoT provide incorrect answers, while RM-PoT gives the correct answer.

## 6 CONCLUSION

In this paper, we present RM-PoT++, a framework that advances mathematical reasoning in LLMs through domain-aware few-shot learning, problem reformulation, and Program of Thoughts. By retrieving semantically aligned examples from a pre-constructed question bank, our method primes LLMs to interpret problems consistently, while reformulation and code generation mitigate structural ambiguities and computational errors. Experiments across datasets demonstrate statistically significant accuracy gains and robustness to linguistic variations.

However, the underlying reasons why naive reformulation can enhance performance are still unclear. In some cases, the reformulated problem even exhibits a lower solve rate than the original. Future work could explore more effective ways to reformulate problems and integrate fine-tuning methods to further improve the performance of LLMs. At the same time, this method can be improved in the following ways:

- Generalization: Extend to geometry and calculus by curating domain-specific banks with diagram-to-code mappings.
- Adaptive Retrieval: Dynamically adjust the number of few-shot examples (K) based on problem complexity.
- Human-in-the-Loop: Integrate user feedback to refine the question bank and domain classifier.